\documentclass[sigconf, anonymous=false]{acmart}

\def\BibTeX{{\rm B\kern-.05em{\sc i\kern-.025em b}\kern-.08emT\kern-.1667em\lower.7ex\hbox{E}\kern-.125emX}}
\usepackage{booktabs}    

\newcommand{\etal}{\textit{et al}.}
\newcommand{\ie}{\textit{i}.\textit{e}.}

\copyrightyear{2019}
\acmYear{2019}
\acmConference[ICDSC 2019]{13th International Conference on Distributed Smart Cameras}{September 9--11, 2019}{Trento, Italy}
\acmBooktitle{13th International Conference on Distributed Smart Cameras (ICDSC 2019), September 9--11, 2019, Trento, Italy}
\acmPrice{15.00}
\acmDOI{10.1145/3349801.3349815}
\acmISBN{978-1-4503-7189-6/19/09}

\begin{document}

%
\title[GA for the Optimization of Diffusion Parameters in CBIR]{Genetic Algorithms for the Optimization of Diffusion Parameters in Content-Based Image Retrieval}

%
\author{Federico Magliani}
\email{federico.magliani@studenti.unipr.it}
\orcid{0000-0001-5526-044}
\affiliation{%
  \institution{IMP lab - University of Parma}
  \streetaddress{Parco Area delle Scienze, 181/A}
  \city{Parma}
  \state{Italy}
  \postcode{43124}
}

\author{Laura Sani}
\email{laura.sani@unipr.it}
\affiliation{%
  \institution{SoWide - University of Parma}
  \streetaddress{Parco Area delle Scienze, 181/A}
  \city{Parma}
  \country{Italy}}

\author{Stefano Cagnoni}
\affiliation{%
  \institution{SoWide - University of Parma}
  \streetaddress{Parco Area delle Scienze, 181/A}
  \city{Parma}
  \state{Italy}}
  
\author{Andrea Prati}
\affiliation{%
  \institution{IMP lab - University of Parma}
  \streetaddress{Parco Area delle Scienze, 181/A}
  \city{Parma}
  \state{Italy}}
 
%
\renewcommand{\shortauthors}{Magliani and Sani, et al.}

%
\begin{abstract}
Several computer vision and artificial intelligence projects are nowadays exploiting the manifold data distribution using, e.g., the diffusion process. This approach has produced dramatic improvements on the final performance thanks to the application of such algorithms to the kNN graph. Unfortunately, this recent technique needs a manual configuration of several parameters, thus it is not straightforward to find the best configuration for each dataset. Moreover, the brute-force approach is computationally very demanding when used to optimally set the parameters of the diffusion approach.
We propose to use genetic algorithms to find the optimal setting of all the diffusion parameters with respect to retrieval performance for each different dataset. Our approach is faster than others used as references (brute-force, random-search and PSO). A comparison with these methods has been made on three public image datasets: Oxford5k, Paris6k and Oxford105k.
\end{abstract}

%
%


%

%

%
\maketitle

\section{Introduction}

The advent of manifold representation and graph-based techniques as diffusion approaches has affected several computer vision research fields, such as  Content-Based Image Retrieval (CBIR). This is a computer vision task, tailored for mobile devices, aimed at ranking increasingly the database images (that can be millions or more) based on the similarity to a query. Similarity is a metric that can be calculated between two vectors that represent the images. The task seems simple but poses several challenges. The algorithm needs to be invariant to: image resolution, illumination conditions, viewpoints, and to the presence of distractors as cars, people and trees \cite{magliani2019landmark}.
Furthermore, the method adopted for the retrieval task needs to be precise (\ie, to obtain a good retrieval performance) and fast (\ie, to retrieve the results in as short time as possible). 
Unfortunately, it is not always possible to obtain excellent results in a short time, therefore the final target is finding a trade-off between these two metrics. The use of descriptors from pre-trained CNN has allowed researchers to obtain good results in a very simple manner: simply extracting the features from an intermediate layer and then applying pooling and normalization techniques. Furthermore, different embedding algorithms for improving the results have been proposed in order to make the descriptors more discriminating and invariant to rotation, change of dimension, occlusions, and so on \cite{babenko2015aggregating, kalantidis2016cross, magliani2018accurate, gordo2016deep}.
\par
Recently, Iscen \etal \cite{iscen2017efficient} and Yang \etal \cite{yang2018efficient} outperformed the state of the art on several public image retrieval datasets through the application of the diffusion process to R-MAC descriptors \cite{gordo2017end}. 
The reason for the success of diffusion for retrieval  \cite{zhou2004ranking} is that it permits to find more neighbors that are close to the query using the manifold representation, than using the Euclidean one.
Although the diffusion improves retrieval results, it requires a long time to create the kNN graph necessary for the diffusion application. To solve this issue we follow the technique proposed by Magliani \etal \cite{magliani2019efficient}, that proposes a method for effective and efficient creation of an approximate kNN graph suitable for the application of the diffusion approach. On this graph it is possible to obtain the same or better retrieval performance after diffusion than using a brute-force approach, requiring a shorter computation time.

As previously said, the diffusion process works well on this task, but it requires the configuration of several parameters in order to obtain the best retrieval performance for each dataset. Some of them are: the number of walks to execute and the number of neighbors in the graph and the number of database images to consider for the random walk process. 
Currently, the configuration of these parameters is obtained through an extensive testing of several different configurations. As an alternative, a brute-force approach could be applied but it is unfeasible due to the huge time necessary to test all possible combinations of the different parameters.

In this paper, we propose to use genetic algorithms to find an optimal configuration of the parameters of the diffusion approach applied to several CBIR datasets. Besides that, the execution of the diffusion process with the correct configuration allows yields very interesting results on several public image datasets, outperforming the state of the art.

The main contributions of this paper are:
\begin{itemize}
    \item the use of genetic algorithms for tuning the diffusion parameters;
    \item the comparison with other different optimization methods which can solve the above problem;
    \item a test of the optimization methods on several public image datasets.
\end{itemize}

The paper is structured as follows. Section 2 introduces the general techniques used in the state of the art. Section 3 describes in detail the graphs and the diffusion mechanism. Section 4 describes the proposed approach. Section 5 reports the experimental results on three public datasets: Oxford5k, Paris6k and Oxford105k. Finally, some concluding remarks are reported.

\section{Related work}
The setting of algorithm parameters has a relevant impact on the performance of machine learning methods. Finding an optimal parameter configuration can be treated as a search problem, aimed at maximizing the quality of a machine learning model, according to some performance metrics (e.g., accuracy). 

One of the main challenges of parameter setting optimization is given by the
complex interactions between the parameters. Configuring the parameters individually may lead to suboptimal results, whereas trying all different combinations
is often impossible due to the curse of dimensionality.

Parameter optimization algorithms can be grouped into two main classes~\cite{eiben1999parameter,ugolotti2019can}:
\begin{itemize}	
	\item  \textit{Parameter tuning}:  the  parameter values are chosen offline and then the algorithm is run using those values, which
	do  not change anymore during execution. This is the case of interest for this paper;
	\item   \textit{Parameter control}:   the
	parameter values  may vary  during the execution,  according to  a strategy
	that  depends on  the  results that  are  being achieved~\cite{karafotias2015parameter}.	
\end{itemize}

The importance of parameter tuning has been frequently addressed in the last years~\cite{montero2018tuners,sipper2018investigating}. Several algorithms for parameter tuning have been 
proposed~\cite{hoos2011automated, bergstra2011algorithms, falkner2018bohb}, among which the simplest 
strategies are grid search and random search. 
In~\cite{bergstra2012random}, the authors compare the performance of neural networks whose hyperparameters have been configured using grid search and random search. They show that random search is more efficient than grid search and able to find models that are as good or better requiring much less computation time. Random search performs better especially when only few hyperparameters affect the final performance of the machine learning algorithm. In this case, grid search allocates too many trials to the exploration of dimensions that do not matter, suffering from poor coverage of dimensions that are important.

When the search space is non-continuous, high-dimensional, non-convex or multi-modal, local search methods are consistently outperformed by stochastic optimization algorithms~\cite{grefenstette1986optimization}. Metaheuristics are general-purpose stochastic procedures designed to solve complex optimization problems~\cite{glover2006handbook,Engelbrecht:2007:CII:1557464}. 
These optimization algorithms are non-deterministic and approximate, i.e., they do not always guarantee that they find the optimal solution, but they can find a good one in reasonable time. Metaheuristics require no particular knowledge about the problem structure other than the objective function itself, when defined, or a sampling of it~\cite{mesejo2016survey}. The main objective of metaheuristics is to achieve a trade-off between diversification (exploration) and intensification (exploitation). Diversification implies generating diverse solutions to explore the search space on a global scale, while exploitation implies focusing the search onto a local region where good solutions have been found. An overview of the main proofs of convergence of metaheuristics to optimal solutions can be found in~\cite{Gutjahr2010}.
\\
Metaheuristics include: 
\begin{itemize}
	\item \textit{Population-based methods}, in which the search process can be seen as the evolution in (discrete) time of a set of points (population of solutions) in the solution space (e.g., evolutionary algorithms~\cite{back1993overview} and particle swarm optimization~\cite{poli2007particle});
	\item \textit{Trajectory methods}, in which the search process describes a trajectory in the search space and can be seen as the evolution in (discrete) time of a discrete dynamical system (e.g., simulated annealing~\cite{kirkpatrick1983optimization});
	\item \textit{Memetic algorithms}, which are hybrid global/local search methods in which a local improvement procedure is combined with a population-based algorithm (e.g., scatter search~\cite{glover2003scatter}).
\end{itemize}

In particular, evolutionary computing 
has been very successful in solving hard, multi-modal, multi-dimensional problems in many different tasks (e.g., parameter tuning~\cite{Rasku2019}). When the dimension of the search space is large, evolutionary computing allows one to perform an efficient directed search, taking inspiration from biological evolution to guide the search~\cite{Eiben:2015:IEC:2810085}. 
In~\cite{konstantinov2019comparative}, the authors present an experimental comparison of evolutionary algorithms and random search algorithms to solve the problem of the optimal control of mobile robots, showing that evolutionary algorithms can find better solutions with the same number of fitness function calculations.

Genetic algorithms (GAs) are evolutionary algorithms inspired by the process of natural selection (survival of the fittest, crossover, mutation, etc.)~\cite{Goldberg:1989:GAS:534133} commonly used to solve optimization problems. In this paper we use a genetic algorithm to optimize the diffusion process, which is a promising approach for image retrieval whose performance depends on the setting of several parameters over different ranges.

\section{Graphs and diffusion}


The k-Nearest Neighbor (kNN) graph is an undirected graph $G$ denoted by $G(V, E)$, where $V$ is the set of nodes $V=\left\{(v_1, v_2, \dots , v_n)\right\}$ and $E$ represents the set of edges $E=\left\{(e_1, e_2, \dots , e_n)\right\}$. 
The nodes represent the dataset images, while the edges are the connections between the nodes. The edges are weighted and these weights determine how much the connected images are similar: the larger the weight, the more similar the two images. 

More formally, starting from a dataset $\mathcal{D} = \{d_1, \dots , d_N\}$, composed by $N$ images, and a similarity measure $\theta : \mathcal{D} \times \mathcal{D} \rightarrow \mathbb{R}$, it is possible to construct the kNN graph for $\mathcal{D}$. It contains edges between nodes $i$ and $j$ whose value is given by the similarity measure $\theta(d_i, d_j) = \theta(d_j, d_i)$. The similarity measure adopted can change depending on the topic. In our case, the cosine similarity is used, so the similarity is calculated through the application of the dot product between the image descriptors.

\subsection{Approximate kNN graph creation}

The creation of the kNN graph is an operation that usually requires much computation time.
The approach that is used more frequently is brute-force, which consists in the connection of each node to all the others. 
In order to reduce computation time and resources, an approximate graph creation method can be used.
There are different methods for constructing the approximate kNN graph. The main strategies are: methods following the divide and conquer strategy, and methods using a local search strategy (e.g., NN-descent \cite{dong2011efficient}). The divide-and-conquer strategy is composed by: the subdivision of the dataset in subsamples (\textit{divide}) and the brute-fore creation of the graphs for all the elements of the subsample (\textit{conquer}).

We follow the idea of Magliani \etal \cite{magliani2019efficient} that exploits the LSH (Locality Sensitive Hashing)  \cite{indyk1998approximate} to approximately split the elements in several buckets using the hash table mechanism. This method can reduce the time required for the creation of the kNN graph, maintaining or, in some cases, improving the final retrieval performance obtained after the diffusion application.

\subsection{Diffusion}

\begin{figure}
\includegraphics[width=0.39\textwidth]{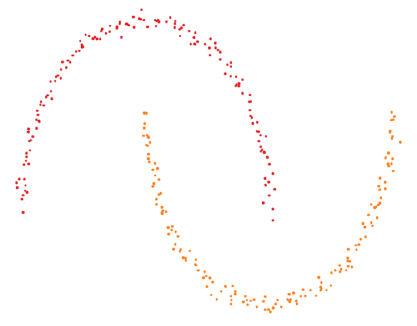}
\caption{Two data distributions (red and orange dots). In this case, the application of an Euclidean metric as the $l_2$ distance does not achieve the best performance. With the diffusion application, that exploits the manifold distribution, better results can be achieved. Best viewed in color.}
\label{fig:manifold_distribution}
\end{figure}

Fig. \ref{fig:manifold_distribution} shows two exemplar data distributions where the diffusion approach is capable to improve the final retrieval performance.

The diffusion is usually applied starting from the query point with the objective to find the neighbors, i.e. images which are the most similar to the query. As mentioned before, the diffusion can be applied only to a kNN graph, that is created based on the database images. The graph is mandatory because it helps to establish the best path from the query to the database points. It is also possible to exploit the similarities between the images (nodes on the graph) in order to find the best path from the database images to the query point on the graph. Indirectly, the nodes crossed on the graph to reach the query represent the neighbors of the query itself, finding which is the objective of the image retrieval task.
The path to follow on the graph is chosen through the application of the random walk process in several iterations. The wrong paths are discarded exploiting the weights of the edges of the kNN graph, which indicate the similarity between two nodes: the greater the weight, the more similar the two nodes.
Mathematically, the entire process is represented by a system of equation $A x = b$, where $A$ is the affinity matrix of the database images (mathematical representation of the graph), $b$ represents the query vector and $x$ is the solution of the system (ranking vector).

\section{Genetic Algorithms for diffusion parameters optimization}
The diffusion process is regulated by several parameters, which can be optimized to improve the retrieval performance. 
\\
In this section we present the diffusion parameters and propose a genetic algorithm for diffusion parameter tuning.

\subsection{Diffusion parameters}
The diffusion approach consists in the resolution of the following equation system: $A x = b$.
The diffusion applied in this paper is similar to the Google PageRank algorithm \cite{page1999pagerank} where a graph is solved by using diffusion iteratively. To achieve this result, the affinity matrix $A$ is modified as follows: $ A = A - \alpha * A$, where $\alpha$ represents the damping factor used in the Google algorithm to adjust the connections between nodes. In their case, the best value for this parameter is set to 0.85, which is obtained after executing many experiments. In our case, this parameter is a real value in the interval $[0,1]$.
Moreover, the elements of the sparse affinity matrix can be raised to power by a factor $\beta$ ($a_{ij} = a_{ij}^\beta$, where $a_{ij} \in A$) in order to remove useless neighbors, similarly to the power iteration method \cite{mises1929praktische} used for the resolution of the PageRank problem. The same reasoning can also be applied to the query vector $b_i = b_i ^ \gamma$, where $b_i \in b$.
The other parameters to optimize are: i) $k_s$, that is the number of steps to execute on the graph during the random walk process; ii) $k$, that is the number of neighbors to find; iii) the maximum number of iterations allowed for the algorithm to converge to the equation system solution ($iterations$); iv) the number of database elements to be used during the application of the diffusion  ($trunc$).

\subsection{Genetic algorithm}
The diffusion parameters have been tuned using a genetic algorithm. Each individual corresponds to a specific setting of diffusion parameters and is represented by a string of seven values, corresponding to the seven parameters. 
The values have been set in the following ranges: $\alpha \in \{0, 1\}~(float)$, $\beta \in \{1, 10\}~(int)$, $\gamma \in \{1, 10\}~(int)$, $k_s \in \{20, 100\}~(int)$, $k \in \{5, 40\}~(int)$, $iterations \in \{10, 30\}~(int)$, $trunc \in \{3000, dataset~size\}~(int)$.

The fitness function to be maximized corresponds to the mean Average Precision (mAP) 
obtained by the diffusion process in the retrieval phase.
It identifies how many elements of an image dataset, on average, are found which are relevant to the query image.  In order to compare a query image with the dataset images, the Euclidean distance is employed.

The initial population, of size $Pop$, is obtained by generating random individuals according to the constraints on the parameter ranges. During the selection operation, each individual is replaced by the best of three individuals extracted randomly from the current generation (tournament selection). The selected individuals are crossed with a probability $CxPb$, generating new individuals by means of a single-point crossover. An individual is mutated with a probability $MutPb$, while each gene is mutated with a probability $IndPb$. The population is then entirely replaced by the offspring (generational GA).
The evolutionary process is iterated for $Gen$ generations.
%

A buffer has been introduced to store the best individuals (those leading to the largest mAP) found during the evolutionary process, and their corresponding fitness (mAP) values.
Thus, at the end of the run, the best parameter setting can be found not only among the individuals of the last population, but also among the best ones found during the whole evolutionary process, which are stored in the buffer.

The genetic algorithm has been implemented using DEAP\footnote{\url{https://deap.readthedocs.io/en/master/}} (Distributed Evolutionary Algorithms in Python)~\cite{fortin2012deap}, an evolutionary computation framework for rapid prototyping and testing.

\section{Experimental results} \label{ref:results}
In this section we illustrate the experimental results we have obtained on three public datasets: Oxford5k, Paris6k and Oxford105k. 
\\
Mean Average Precision (mAP) is used on all image datasets to evaluate the accuracy in the retrieval phase.

The results of the GA optimization are compared to the results obtained by other commonly used techniques for parameter tuning.


\subsection{Datasets}
To evaluate the optimization of the diffusion parameters, the experiments are applied on several CBIR public image datasets:
\begin{itemize}
\item \textbf{Oxford5k} \cite{Philbin07} contains 5063 images belonging to 11 classes. 
\item \textbf{Paris6k} \cite{Philbin08} contains 6412 images belonging to 12 classes.
\item \textbf{Flickr1M} \cite{huiskes2008mir} contains 1 million Flickr images used for large scale evaluation. The images are divided in multiple classes and are not specifically selected for the image retrieval task. 
\end{itemize}

With the addition of 100k images of Flickr1M it is possible to create the dataset \textbf{Oxford105k}.


\subsection{Results on Oxford5k}

Different experiments have been executed on the Oxford5k dataset. In order to find the best configuration of the diffusion parameters, several combination of genetic algorithm parameters have been tested.

\begin{table}[ht]
\centering
\setlength{\tabcolsep}{3pt}
    \begin{tabular}{c | c | c | c | c | c }
    \toprule
    \textbf{Gen} &  \textbf{Pop} & \textbf{CxPb} & \textbf{MutPb} & \textbf{IndPb} & \textbf{mAP}  \\ \midrule
	10 & 50 & 0.5 & 0.2  & 0.1 &  94.31\% \\ 
	20  & 50 & 0.5 & 0.2 & 0.1 & 94.31\% \\ 
	50 & 50 & 0.5 &  0.2 & 0.1 & \textbf{94.40\%} \\ 
	100  & 50 & 0.5 & 0.2 & 0.1 &  \textbf{94.40\%} \\ \bottomrule
   \end{tabular}
        \caption{Results on Oxford5k varying the values of number of generations ($Gen$).}

        \label{results_oxf5k_generation}
\end{table}

\begin{table}[ht]
\centering
\setlength{\tabcolsep}{3pt}
    \begin{tabular}{c | c | c | c | c | c }
    \toprule
    \textbf{Gen} &  \textbf{Pop} & \textbf{CxPb} & \textbf{MutPb} & \textbf{IndPb} & \textbf{mAP}  \\ \midrule
    50  & 10 & 0.5 & 0.2 & 0.1 & 94.32\% \\ 
	50 & 20 & 0.5 &  0.2 & 0.1 & 94.16\% \\ 
	50 & 50 & 0.5 &  0.2 & 0.1 & \textbf{94.40\%} \\ 
	50 & 100 & 0.5 &  0.2 & 0.1 & 94.36\% \\  \bottomrule
   \end{tabular}
        \caption{Results on Oxford5k varying the values of population size ($Pop$).}

        \label{results_oxf5k_population}
\end{table}

\begin{table}[ht]
\centering
\setlength{\tabcolsep}{3pt}
    \begin{tabular}{c | c | c | c | c | c }
    \toprule
    \textbf{Gen} &  \textbf{Pop} & \textbf{CxPb} & \textbf{MutPb} & \textbf{IndPb} & \textbf{mAP}  \\ \midrule
    50  & 50 & 0.1 & 0.2 & 0.1 & 93.73\% \\ 
	50 & 50 & 0.3 &  0.2 & 0.1 & \textbf{94.41\%} \\ 
	50 & 50 & 0.5 &  0.2 & 0.1 & 94.40\% \\ 
	50 & 50 & 0.8 &  0.2 & 0.1 & 94.36\% \\ 
	50 & 50 & 1.0 &  0.2 & 0.1 & 94.34\% \\ \bottomrule
   \end{tabular}
        \caption{Results on Oxford5k varying the values of crossover probability ($CxPb$).}

        \label{results_oxf5k_crossover}
\end{table}

\begin{table}[ht]
\centering
\setlength{\tabcolsep}{3pt}
    \begin{tabular}{c | c | c | c | c | c }
    \toprule
    \textbf{Gen} &  \textbf{Pop} & \textbf{CxPb} & \textbf{MutPb} & \textbf{IndPb} & \textbf{mAP}  \\ \midrule
    50 & 50 & 0.3 & 0.1  & 0.1 &  93.73\% \\
    50 & 50 & 0.3 & 0.2  & 0.1 &  \textbf{94.41\%} \\
    50  & 50 & 0.3 & 0.3 & 0.1 & \textbf{94.41\%} \\
	50  & 50 & 0.3 & 0.4 & 0.1 & 94.32\% \\  
	50  & 50 & 0.3 & 0.5 & 0.1 & 94.31\% \\ \bottomrule
   \end{tabular}
        \caption{Results on Oxford5k varying the values of mutation probability ($MutPb$).}

        \label{results_oxf5k_mutpb}
\end{table}

\begin{table}[ht]
\centering
\setlength{\tabcolsep}{3pt}
    \begin{tabular}{c | c | c | c | c | c }
    \toprule
    \textbf{Gen} &  \textbf{Pop} & \textbf{CxPb} & \textbf{MutPb} & \textbf{IndPb} & \textbf{mAP}  \\ \midrule
    50  & 50 & 0.3 & 0.2 & 0.1 & \textbf{94.41\%} \\ 
	50 & 50 & 0.3 &  0.2 & 0.3 & 94.40\% \\ 
	50 & 50 & 0.3 &  0.2 & 0.5 & 94.27\% \\ 
	50 & 50 & 0.3 &  0.2 & 0.8 & 94.23\% \\
	50 & 50 & 0.3 &  0.2 & 1.0 & 94.22\% \\ \bottomrule
   \end{tabular}
        \caption{Results on Oxford5k varying the values of mutation probability for each gene ($IndPb$).}

        \label{results_oxf5k_indpb}
\end{table}

Tables \ref{results_oxf5k_generation}-\ref{results_oxf5k_indpb} report the results obtained on Oxford5k by varying one parameter of the genetic algorithm at a time. 
Starting from a standard configuration of the GA ($CxPb=0.5$, $MutPb=0.2$ and $IndPb=0.1$), the number of generations and the population size have been varied from $10$ to $100$, considering a maximum budget of $5000$ fitness computations. The best configurations, as shown in Table~\ref{results_oxf5k_generation} and~\ref{results_oxf5k_population}, correspond to the largest numbers of fitness computations ($Gen=50$, $Pop=50$ and $Gen=100$, $Pop=50$). Since these configurations lead to the same mAP ($94.40\%$), the remaining parameters of the GA have been varied starting from the configuration which is fastest to compute ($Gen=50$, $Pop=50$).

Table~\ref{results_oxf5k_crossover} shows that the precision reaches its highest value for a crossover probability ($CxPb$) of $0.3$ ($94.41\%$). Regarding the mutation probability ($MutPb$), the best results have been achieved with values $0.2$ ($94.41\%$) and $0.3$ ($94.41\%$), as shown in Table~\ref{results_oxf5k_mutpb}. Considering the mutation probability for each gene ($IndPb$), the highest precision has been achieved with value $0.1$ ($94.41\%$).

Therefore, as shown in Table~\ref{results_oxf5k_indpb}, the best set of parameters for the genetic algorithm thus obtained is: $Gen$= $50$, $Pop$= $50$, $CxPb$= $0.3$, $MutPb$= $0.2$, $IndPb$= $0.1$. The corresponding configuration obtained for the diffusion parameters is: $\alpha = 0.97$, $\beta = 3$, $\gamma = 2$, $k_s = 53$, $k = 9$, $iterations = 10$, $trunc = 4136$.
%

After this preliminary analysis, another set of experiments has been performed. The number of generations has been increased in order to check the convergence status of the GA, obtaining a further improvement in the mAP. It is to be noticed that this set of experiments is less structured than the previous one, due to the longer computation time. 
The best set of GA parameters thus obtained ($mAP=94.44\%$) is: $100$ generations, population size equal to $50$, crossover probability set to $0.3$, mutation probability to $0.2$ and mutation probability for each gene to $0.1$. The corresponding configuration obtained for the diffusion parameters is: $\alpha = 0.97$, $\beta = 3$, $\gamma = 1$, $k_s = 95$, $k = 7$, $iterations = 10$, $trunc = 3046$. Given the stochastic nature of the GA, five independent runs of the algorithm have been executed to assess how repeatable the results are (avg = 94.39\%, stdev = 0.038, max = 94.44\%, min = 94.34\%).

\begin{table}[ht]
\centering
\setlength{\tabcolsep}{4pt}
    \begin{tabular}{c | c | c | c }
    \toprule
    \textbf{Method}  & \textbf{Fitness comp.} & \textbf{Time} & \textbf{mAP}  \\ \midrule
	genetic algorithms  & \textbf{5000} & \textbf{17695 s} & \textbf{94.44\%} \\ 
	PSO \cite{poli2007particle} & 5000 & 27767 s & 94.30\% \\
	random search \cite{bergstra2012random} & 20000  & 27045 s & 93.67\% \\
	grid search & 200000  & 1036800 s & 94.43\% \\ 
	manual configuration \cite{magliani2019efficient} & 1 & 2 s & 90.95\% \\ \bottomrule
   \end{tabular}
        \caption{Comparison of different approaches to the optimization of the diffusion parameters on Oxford5k in terms of mAP, time and number of fitness computations.}

        \label{results_oxf5k}
\end{table}

Table \ref{results_oxf5k} reports the results of different optimization techniques applied on the diffusion process. For each technique the table shows the result of the best configuration found. The results have been compared in terms of mAP, running time and number of fitness computations.
\\
The random search \cite{bergstra2012random}
has sampled, in this case, 20k configurations using uniform distribution for all the parameters to test.
\\
The Particle Swarm Optimization \cite{poli2007particle} has been executed using the same number of fitness computation of the GA (population of $50$ particles, $100$ iterations).
Moreover, the minimum speed is set to $-0.50$ and the maximum speed to $0.50$.
\\
The grid search has been performed over 200k different parameter setting. Given the large number of fitness computations it can be seen as a brute-force strategy.
\\
``Manual configuration" means that the configuration of the parameters of the diffusion mechanism was taken from the literature.

The ``manual configuration" technique obviously requires less time than the other methods, but it obtains the worst final results. The genetic algorithms achieve an excellent result in much shorter time than the others. It is to be noticed that, in all the previous experiments, the GA has performed better than manual configuration and random search.
Thus, only the manual configuration and the GAs have been tested on the other datasets.
The results of PSO are comparable, but the computation time required to perform the same number of fitness computations as the GA is longer.




\subsection{Results on Paris6k}

\begin{table}[ht]
\centering
\setlength{\tabcolsep}{4pt}
    \begin{tabular}{c | c | c  }
    \toprule
    \textbf{Method}  &  \textbf{Time} & \textbf{mAP} \\ \midrule
	genetic algorithms & 18787 s  & \textbf{97.32\%}  \\ 
	manual configuration \cite{magliani2019efficient} & 4 s   & 97.01\% \\ \bottomrule
   \end{tabular}
        \caption{Comparison of different approaches for the optimization of the diffusion parameters on Paris6k.}

        \label{results_par6k}
\end{table}
\vspace{-5mm}

Table \ref{results_par6k} reports the results of different optimization methods applied on Paris6k.
The best result (97.32\%) has been obtained with the following GA configuration: 
$Gen= 100$, $Pop= 50$, $CxPb= 0.5$, $MutPb= 0.2$, $IndPb= 0.1$.
The final configuration of the diffusion parameters is:
$\alpha = 0.87$, $\beta = 1$, $\gamma = 2$, $k_s = 40$, $k = 11$, $iterations = 10$, $trunc = 3761$.

As in the previous dataset, the GAs need more computation time than the ``manual configuration", but they improve the final performance of the diffusion process for retrieval.

\subsection{Results on Oxford105k}
Given the large dimension of Oxford105k dataset, the ranges of parameters $k_s$ and $k$ have been extended to $\{20, 250\}~(int)$ and $\{5, 100\}~(int)$, respectively.

Table \ref{results_oxf105k} reports the results of different optimization methods applied on Oxford105k.

\begin{table}[ht]
\centering
\setlength{\tabcolsep}{4pt}
    \begin{tabular}{c | c | c }
    \toprule
    \textbf{Method}  &  \textbf{Time} & \textbf{mAP}  \\ \midrule
	genetic algorithms  & 63911 s & \textbf{94.20\%} \\ 
	manual configuration \cite{magliani2019efficient} & 13 s  & 92.50\% \\ \bottomrule
   \end{tabular}
        \caption{Comparison of different approaches for the optimization of the diffusion parameters on Oxford105k.}

        \label{results_oxf105k}
\end{table}

The best result (94.20\%) is obtained with the following GA configuration: 
$Gen= 100$, $Pop= 50$, $CxPb= 0.5$, $MutPb= 0.2$, $IndPb= 0.1$.
The final configuration of the diffusion parameters is:
$\alpha = 0.97$, $\beta = 2$, $\gamma = 1$, $k_s = 68$, $k = 7$, $iterations = 10$, $trunc = 18353$.

The ``manual configuration" is faster than the GAs, but the final performance is very different: the GAs obtain 94.20\% while the "manual configuration" achieves only 92.50\%.

\section{Conclusions}

In this paper we propose to use genetic algorithms for searching the optimal configuration of the diffusion parameters using kNN graphs within the field of Context-Based Image Retrieval (CBIR). By applying genetic algorithms to this optimization problem, a better set of parameters has been obtained, resulting in a higher precision of the retrieval when applied to several public image datasets. Comparing our method with other techniques, as random search, grid search and PSO, our optimization approach is faster and obtains the same or better retrieval results.
It should be noticed that, despite our objective to find a common set of parameters for all the datasets, it turns out that the optimization needs to be tailored on a specific dataset in order to achieve the best result.


Finally, we will further study the dependence of the GA on its parameters, to improve its effectiveness using Meta-EAs, methods that tune the parameters of evolutionary algorithms to optimize their performance.

\section*{Acknowledgments}
The work by Federico Magliani and Laura Sani was funded by Regione Emilia Romagna within
the ``Piano triennale alte competenze per la ricerca, il trasferimento tecnologico e l'imprenditorialit\`a" framework. The work of Laura Sani was also co-funded by Infor srl.

%
\bibliographystyle{ACM-Reference-Format}
\bibliography{sample-base}

%









\end{document}